\RequirePackage[T1]{fontenc}
\documentclass[conference,letterpaper]{IEEEtran}
\IEEEoverridecommandlockouts
\usepackage{cite}
\usepackage{amsmath,amssymb,amsfonts}
\usepackage{graphicx}
\usepackage{textcomp}
\usepackage{xcolor}
\def\BibTeX{{\rm B\kern-.05em{\sc i\kern-.025em b}\kern-.08em
    T\kern-.1667em\lower.7ex\hbox{E}\kern-.125emX}}

\usepackage{booktabs}
\usepackage{url}
\usepackage{tikz}
\usetikzlibrary{arrows.meta,positioning}
\usepackage[hidelinks,pdfversion=1.7]{hyperref}

\newcommand{\Pavg}{P_{\mathrm{avg}}}
\newcommand{\EOpp}{\mathrm{EOpp}}
\newcommand{\R}{\mathbb{R}}

\begin{document}

\title{
FairCompressAgent: An Agentic Framework for Fairness-Aware Model Compression for FPGA Deployment
}

\author{\IEEEauthorblockN{Yuanbo Guo}
\IEEEauthorblockA{\textit{Department of Computer Science and Engineering} \\
\textit{University of Notre Dame}\\
Notre Dame, USA \\
yguo6@nd.edu}
\and
\IEEEauthorblockN{Yiyu Shi}
\IEEEauthorblockA{\textit{Department of Computer Science and Engineering} \\
\textit{University of Notre Dame}\\
Notre Dame, USA \\
yshi4@nd.edu}
}

\maketitle

\begin{abstract}
Fairness-aware model compression requires selecting methods and configurations that balance accuracy, fairness, and deployment cost.
These decisions become more difficult when compression methods are composed or the user's requirements change.
In this paper, we propose FairCompressAgent (FCA), an agentic framework that integrates fairness-aware pruning, incremental quantization, and sparse low-rank factorization through a common operator interface.
A language-model planner uses model profiles and measured outcomes to select compression configurations, while an execution layer performs compression, fine-tuning, evaluation, and constraint-based selection.
FCA also supports requirement updates and reports the remaining violation when a request cannot be satisfied.
Experiments on Fitzpatrick-17k with VGG-11 compare four search methods over 40 measured configurations.
Under the accuracy-constrained request, FCA selects a compressed model with 59.54\% less inference tensor storage, while validation average precision increases from 0.5141 to 0.5233 and equalized opportunity (EOpp) decreases from 0.2251 to 0.2168.
It reaches the same final selection as one-shot planning with 7.33 versus 12 candidate evaluations on average, under their respective stopping policies.
Repeated fine-tuning, held-out testing, and online requirement updates characterize the stability and interactive use of this compression workflow.
The results demonstrate how measured feedback and explicit constraints support the selection and interactive refinement of fairness-aware compression configurations.
\end{abstract}

\begin{IEEEkeywords}
deep learning, fairness, hardware efficiency, AI agent, FPGA.
\end{IEEEkeywords}

\section{Introduction}\label{sec:introduction}

As deep learning (DL) becomes increasingly adopted in different areas of daily life, its applications raise a variety of issues that need to be addressed.
Among these concerns, fairness and efficiency are often treated as separate problems.
To begin with, fairness concerns arise from discrimination in prediction outcomes, such as employment discrimination~\cite{dastin2022amazon,ferrara2024fairness}, as well as disparities in prediction quality, such as healthcare inequity~\cite{ferrara2024fairness,obermeyer2019dissecting}.
DL models can influence people's decisions, so applications supported by unfair models may contribute to systematic discrimination and disadvantage particular demographic groups.
To deal with fairness issues, previous works majorly focus on algorithm optimization~\cite{xu2026group,xu2025incorporating,chiu2023toward}.
Meanwhile, model compression is one of the most useful approaches for deploying DL models locally with limited hardware resources~\cite{han2015deep}, including FPGA platforms.
For instance, pruning~\cite{han2015learning,luo2017thinet} and quantization~\cite{jacob2018quantization,zhou2016dorefa} are two widely used techniques for model compression.
Conventionally, however, compression decisions mainly balance prediction quality and computational cost, without explicitly considering unequal performance across demographic groups.
~\cite{hooker2020characterising} also pointed out that model compression could disproportionately affect bias even when accuracy performance changes little.

On top of that, FPGA deployment is particularly relevant to model compression.
Available compute resources and memory bandwidth jointly constrain accelerator design, while on-chip storage influences data reuse and transfers to external memory~\cite{zhang2015optimizing}.
Low-precision designs further connect numerical representation with parallelism and dataflow~\cite{umuroglu2017finn}.
Different compression methods address these constraints in different ways.
For example, pruning changes connectivity of neurons, while quantization restricts how weights are represented.
The resulting representation determines which storage and execution opportunities a target design can exploit.
Choosing a suitable compression configuration therefore requires considering operator compatibility and resource requirements together with inference performance in terms of both accuracy and fairness~\cite{guo2024hardware}.

Recent works start to explore the relationship between DL model fairness and hardware platforms~\cite{mugdho2025fairxbar,sheng2024data,qin2024fl,sheng2022larger}.
Specifically, fairness-aware compression methods have been introduced to seek fairness improvements during compression, creating a trade-off among accuracy, fairness, and efficiency.
FairPrune~\cite{wu2022fairprune} addresses this problem through group-sensitive pruning, and FairQuantize~\cite{guo2024fairquantize} extends the idea to incremental weight quantization.
In addition, sparse low-rank factorization offers another way to reduce model complexity~\cite{swaminathan2020sparse}, and FairLRF adapts it to group-sensitive compression~\cite{guo2025fairlrf}.
There have also been a few works following these paths~\cite{kong2024achieving}.
While all these methods provide different ways to address fairness through compression, choosing among them still requires decisions about compression strength, fine-tuning, layer coverage, and method order.

Joint compression search and language-model-assisted optimization provide foundations for coordinating these decisions.
APQ jointly searches architecture, pruning, and quantization policies~\cite{wang2020apq}, while LLAMBO investigates language models within Bayesian optimization~\cite{liu2024large}.
FairAgent uses language-model agents to support fairness-aware machine learning~\cite{dai2025fairagent}, and ProfilingAgent uses profiling results to guide model optimization~\cite{jafari2025profilingagent}.
Applying these ideas to fairness-aware compression requires a common interface for compatible operators and explicit criteria for evaluating accuracy, fairness, and efficiency.

\begin{figure}[t]
\centering
\begin{tikzpicture}[
 node distance=2.5mm,
 box/.style={draw,rounded corners=1pt,align=center,text width=0.81\columnwidth,inner sep=4pt,font=\footnotesize},
 flow/.style={-{Latex[length=1.5mm]},semithick}
]
\node[box,fill=black!4] (input) {Specify a model, annotated data, and requirements\\Accuracy, fairness, storage, and search budget};
\node[box,below=of input,fill=blue!5] (search) {FCA evaluates compression configurations\\Compare measured trade-offs under the requirements};
\node[box,below=of search] (inspect) {Inspect the selected model and report\\Review performance, resource use, and unmet constraints};
\node[box,below=of inspect,fill=black!4] (decision) {Accept the result or revise the requirements\\Continue with the accumulated observations};
\draw[flow] (input) -- (search);
\draw[flow] (search) -- (inspect);
\draw[flow] (inspect) -- (decision);
\draw[flow] (decision.east) -- ++(4mm,0) |- (search.east);
\end{tikzpicture}
\caption{
User interaction with FCA.
The user supplies the task and requirements, inspects measured results, and can revise the bounds to continue with the accumulated evaluation records.
}
\label{fig:overview}
\end{figure}
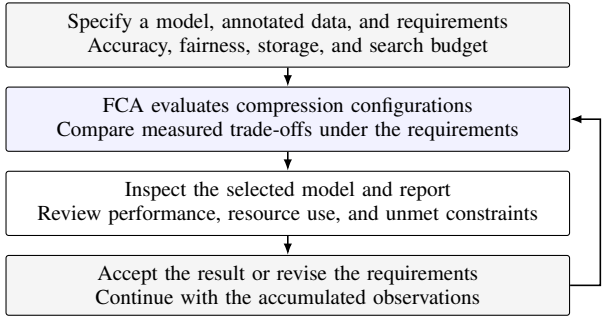

To integrate these capabilities into comprehensive workflows, we propose FairCompressAgent (FCA), an agentic framework for selecting and composing fairness-aware compression operators under user-defined requirements.
FCA uses model profiles and evaluation results to guide compression choices and allows the user to revise requirements during the search.
Figure~\ref{fig:overview} summarizes the interaction between the user and FCA.

The main contributions of this paper are as follows:
\begin{itemize}
\item We formulate fairness-aware compression as constrained configuration selection and organize pruning, quantization, and sparse low-rank factorization through a common perturbation view and executable interface.
\item We develop an agent architecture for adaptive compression planning, with explicit constraint checks and support for requirement updates.
\item We evaluate the framework using 40 measured configurations, four search methods, repeated fine-tuning, held-out testing, and fresh online execution. FCA selects a model with 59.54\% less inference tensor storage under the accuracy-constrained request and identifies the unmet constraint under a stricter fairness request.
\end{itemize}
The source code for FairCompressAgent is available at \url{https://github.com/guoyb17/FairCompressAgent}.

\section{Problem Formulation}\label{sec:problem}

\subsection{Inputs and Accuracy Metrics}

Consider a classification dataset $\mathcal{D}=\{(x_i,y_i,a_i)\}_{i=1}^{N}$, where $x_i$ is the input, $y_i$ is the class label, and $a_i$ is a sensitive attribute.
We focus on binary sensitive attributes, which divide the annotated samples into groups $g\in\{0,1\}$.
Given a pre-trained model $f(\theta_0)$, our objective is to find a compressed model that satisfies the user's requirements on accuracy, fairness, and efficiency.

Following FairPrune, FairQuantize and FairLRF, we use average precision as the main measure of accuracy performance.
It is the mean of the macro precision values of the two demographic groups.
For a fixed set of $K$ classes, it is defined as
\begin{equation}
 P_g=\frac{1}{K}\sum_{k=1}^{K}\operatorname{Prec}_{g,k},
 \qquad \Pavg=\frac{P_0+P_1}{2},
 \label{eq:utility}
\end{equation}
where $\operatorname{Prec}_{g,k}$ is the one-versus-rest precision of class $k$ in group $g$.
Precision is set to zero when no sample in that group is predicted as class $k$.

\subsection{Fairness and Efficiency Metrics}

Equalized opportunity and equalized odds~\cite{hardt2016equality} have been widely used in previous research regarding fairness.
For equalized opportunity, it is the difference of true negative rates (denoted as EOpp0) or true positive rates (denoted as EOpp1) of different demographic groups within each class of target attributes, followed by sum of them.
For equalized odds (denoted as EOdd), it collects the difference of true positive rates plus difference of false positive rates.
These metrics can be calculated by the following equations:
\begin{equation}
    EOpp0 = \sum_{k=1}^{K}{ \lvert TNR_k^1 - TNR_k^0 \rvert },
\end{equation}
\begin{equation}
    EOpp1 = \sum_{k=1}^{K}{ \lvert TPR_k^1 - TPR_k^0 \rvert },
\end{equation}
\begin{equation}
    EOdd = \sum_{k=1}^{K}{ \lvert TPR_k^1 - TPR_k^0 + FPR_k^1 - FPR_k^0 \rvert }.
\end{equation}
For all of EOpp0, EOpp1, and EOdd, the smaller the better.

These metrics reflect the statistical differences of model performance on different demographic groups from different perspectives.
Given that fairness for positive cases is usually more important in real world scenarios like disease diagnosis, we use EOpp1 as the representation of fairness metrics in this paper, denoted as EOpp below.

A compressed model is represented by $z=(\theta,\rho)$, where $\rho$ specifies its executable form, including factor layers and constrained weights.
We measure storage $S(\rho)$ as the bytes occupied by inference parameter and buffer tensors, excluding optimizer state, training masks, and file headers.
The storage measurement therefore reflects the representation used for inference.
Execution cost additionally depends on how the representation maps to the target architecture.
The present experiments use dense FP32 tensors, including the factors of low-rank layers.

\subsection{Constrained Configuration Selection}

A configuration $\pi=((o_1,\eta_1),\ldots,(o_T,\eta_T))$ specifies compression operators $o_t$, their order, and settings $\eta_t$.
Each configuration starts from the supplied baseline and produces a candidate $z_\pi$.
The three objectives are higher $\Pavg$, lower EOpp, and smaller storage:
\begin{equation}
 \min_{\pi}\bigl(-\Pavg(z_\pi),\ \EOpp(z_\pi),\ S(\rho_\pi)\bigr).
 \label{eq:objectives}
\end{equation}
The user specifies acceptable average precision loss $\epsilon_P$, an EOpp upper bound $d_{\max}$, and, optionally, a storage-ratio upper bound $s_{\max}$:
\begin{align}
 \Pavg(z_\pi)&\geq\Pavg(z_0)-\epsilon_P,\label{eq:utilityconstraint}\\
 \EOpp(z_\pi)&\leq d_{\max},\label{eq:fairnessconstraint}\\
 S(\rho_\pi)/S(\rho_0)&\leq s_{\max}.\label{eq:storageconstraint}
\end{align}
Among feasible observations, the experiments select minimum storage, then maximum $\Pavg$, then minimum EOpp, with candidate identifiers breaking any remaining ties.

\section{Fairness-Aware Compression Operators}\label{sec:operators}

\subsection{A Common Group-Sensitive Score}

Different weights can contribute unequally to the prediction performance of different demographic groups, and second-order sensitivity provides a promising basis for ranking weight perturbations~\cite{lecun1989optimal}.
FairPrune and FairQuantize use this observation to decide which weights to modify, while FairLRF applies a related score to low-rank factors~\cite{wu2022fairprune,guo2024fairquantize,guo2025fairlrf}.
We organize these existing methods as compression operators with a common scoring and evaluation interface.

Let $L_g(\theta)$ be the mean loss on a fixed scoring subset for group $g$.
For a weight perturbation $\Delta$, the Taylor expansion gives
\begin{equation}
 L_g(\theta+\Delta)-L_g(\theta)
 =\nabla L_g^{\top}\Delta+\tfrac12\Delta^{\top}H_g\Delta+R_g(\Delta),
 \label{eq:taylor}
\end{equation}
where $H_g$ is the group-specific Hessian and $R_g$ contains higher-order terms.
The operators use the following diagonal quadratic score:
\begin{equation}
 s_i=\tfrac12\Delta_i^2\left(h_{u,i}-\beta h_{p,i}\right),
 \qquad \beta\geq0,
 \label{eq:score}
\end{equation}
Here $h_{g,i}$ is the $i$th diagonal entry of $H_g$, and $u$ and $p$ are the reference underperforming and better-performing groups.
The experiment fixes these roles as Light and Dark from the baseline comparison throughout search.
A smaller score favors changes with less estimated loss impact on group $u$ relative to group $p$.
The hyper-parameter $\beta$ adjusts the local ranking, while~\eqref{eq:fairnessconstraint} governs the measured fairness of the resulting model.

The local score ranks proposed modifications.
Full evaluation after fine-tuning determines whether the resulting model satisfies the user's requirements.

\subsection{Pruning and Incremental Quantization}

For FairPrune, removing a weight sets it to zero, giving $\Delta_i=-\theta_i$ in~\eqref{eq:score}~\cite{wu2022fairprune}.
The operator selects the required number of lowest-scoring eligible weights within each target tensor.
A cumulative mask keeps these weights at zero during subsequent fine-tuning.

For FairQuantize, the displacement is $\Delta_i=Q(\theta_i)-\theta_i$~\cite{guo2024fairquantize}.
The implementation uses a signed power-of-two mapping,
\begin{equation}
 Q(w)=\begin{cases}
 0,&w=0,\\
 \operatorname{sign}(w)\,2^{\operatorname{round}(\log_2|w|)},&w\neq0.
 \end{cases}
 \label{eq:quantize}
\end{equation}
The squared displacement in~\eqref{eq:score} accounts for the size of each quantization change.
Following incremental network quantization~\cite{zhou2017incremental}, the selected values are frozen while the remaining weights receive fine-tuning updates.
Previously frozen entries remain fixed during subsequent increments.
The resulting power-of-two weights provide a representation that can be mapped to a compatible quantized execution backend.

\subsection{Sparse Low-Rank Factorization}

For a linear layer with $m$ inputs and $n$ outputs, write its weight matrix in input-by-output coordinates as $W\in\R^{m\times n}$.
Truncated singular value decomposition gives
\begin{equation}
 W\approx U_r\Sigma_rV_r^{\top},
 \label{eq:lrf}
\end{equation}
where $r$ is the retained rank.
Sparse low-rank factorization further removes selected factor entries~\cite{swaminathan2020sparse}.
FairLRF~\cite{guo2025fairlrf} determines where this sparsity is applied using group-sensitive scores.
Let $q=\lfloor\alpha r\rfloor$ be the retained prefix of singular directions.
For row $i$ of $U_r$, its tail-removal score is
\begin{equation}
 S_i^U=\sum_{j=q+1}^{r}\tfrac12U_{ij}^{2}
 \left(h_{u,ij}^{U}-\beta h_{p,ij}^{U}\right).
 \label{eq:lrfscore}
\end{equation}
The corresponding expression scores the tail of each column of $V_r^{\top}$.
The lowest-scoring rows and columns receive tail removal according to a sparsification fraction.
Curvature is computed in the factor coordinates with the singular values fixed during scoring.
We use the quadratic tail sum in~\eqref{eq:lrfscore} throughout this study.

After sparsification, singular values are absorbed into one factor, and inference uses two consecutive linear maps with the original bias applied at the output.
Their matrix storage is $r(m+n)$ FP32 elements, compared with $mn$ for the original layer.
The factor masks are preserved during fine-tuning.
The evaluated factorization operator targets fully connected layers.

\subsection{Composition of Operators}

We denote the three operators by $P$, $Q$, and $L$.
Besides individual operators, FCA supports $P\!\rightarrow\!Q$ and $L\!\rightarrow\!Q$.
For $P\!\rightarrow\!Q$, quantization acts on the remaining unpruned weights, and all pruning zeros are preserved.
For $L\!\rightarrow\!Q$, the quantizer operates directly on the executable factors and recomputes curvature after the low-rank transformation and its fine-tuning stage.
Factor sparsity remains fixed during quantization and further fine-tuning.
The executor restores frozen values after each optimizer step and clears their gradients and momentum updates.

Order matters because the same representation property need not survive the reverse composition.
For example, factorizing a quantized matrix generally produces factors outside its original quantization codebook.
Similarly, factorizing a pruned matrix need not preserve its zeros.
These reverse orders are therefore excluded from the present configuration catalogue.

\subsection{Operator Information for Planning}

The planner receives each operator's supported layers, settings, and representation changes.
For a linear layer, two dense factors reduce matrix storage when $r<mn/(m+n)$.
The executor exposes storage for the retained rank, while evaluation supplies average precision and EOpp after fine-tuning.
Pruning masks and quantized values specify constraints for subsequent transformations.
This information lets FCA compare individual methods and compatible compositions under the same selection rule.

\section{FCA Architecture}\label{sec:agent}

\begin{figure*}[t]
\centering
\includegraphics[width=0.75\textwidth]{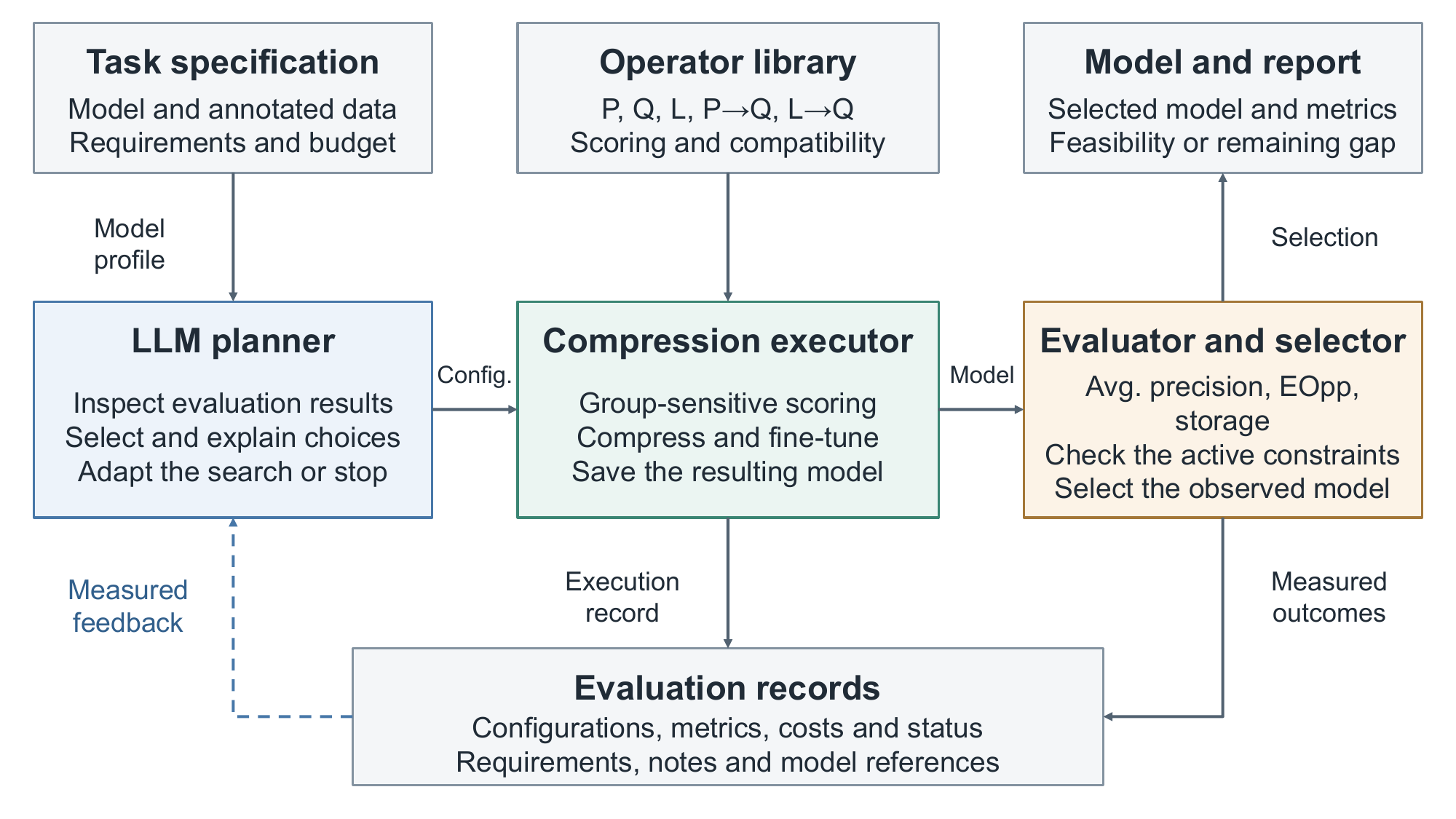}
\caption{Overall architecture of FCA.
The planner selects configurations using the model profile, user requirements, and accumulated evaluation records.
The operator library supplies compatible transformations to the executor, and the evaluator returns measured outcomes to the shared session record.
Constraint-based selection produces the chosen model and report; the feedback path supports subsequent planning and requirement updates.}
\label{fig:framework}
\end{figure*}

\subsection{Planning, Execution, and Evaluation}

Reasoning-and-action agents provide a precedent for interleaving planning with tool feedback~\cite{yao2022react}.
Figure~\ref{fig:framework} presents the three main components of FCA: a language-model planner, a compression executor, and an evaluator.
The planner receives baseline performance, group coverage, layer dimensions, supported configurations, and the user's constraints and search budget.
It selects from a finite catalog of individual operators and compatible compositions.
Each configuration specifies the operator sequence, target layers, compression strength, and group-weighting parameter $\beta$.
The planner combines structural information, such as the relationship between retained rank and tensor storage, with queried evaluation results to select a configuration and explain its choice.

The executor handles model transformations and fine-tuning, while the evaluator measures $\Pavg$, EOpp, group coverage, inference tensor storage, and execution cost.
Typed tool interfaces restrict the planner to supported configurations and operations, including evaluation, comparison, requirement updates, and report export.
Images, individual predictions, and model weights remain local; compact profiles and queried outcomes enter the planner's context.

\subsection{State and Decision Cycle}

The session state at step $t$ consists of the current requirements $\mathcal{C}_t$, remaining budget $b_t$, and evaluation records $E_t$.
Each record stores the configuration, measured metrics, execution cost, completion status, and saved model reference.

Before an evaluation, FCA checks that the requested configuration is supported and that sufficient budget remains.
A repeated request retrieves its existing record.
For a new configuration in online execution, the executor performs compression, fine-tuning, and validation from the supplied baseline.
FCA then updates $E_t$ and $b_t$ and applies the selection rule in Section~\ref{sec:problem} under $\mathcal{C}_t$.
Failed evaluations remain in the record with their status.

\subsection{Measured Feedback and Requirement Updates}

For controlled comparison of search methods, FCA retrieves measured outcomes and execution times from a shared pre-evaluated candidate set.
The planner receives only queried outcomes, and each distinct query contributes its recorded execution time to the accumulated model evaluation time.
This replay protocol keeps the candidate measurements identical across search methods.

When the user revises a requirement, the update tool records the instruction, changes $\mathcal{C}_t$, and reapplies the selection rule to $E_t$ without further model execution.
Each reported selection retains the requirements under which it was made.

\subsection{Stopping and Unmet Constraints}

Each session has limits on distinct evaluations, accumulated model evaluation time, and language-model requests.
The planner may stop earlier when the observed trade-offs are sufficient for its response.

When the observed set contains no feasible candidate, FCA reports the unmet request and identifies the smallest normalized violation for diagnosis:
\begin{equation}
 \begin{split}
 v(z)={}&\left[\frac{\Pavg(z_0)-\Pavg(z)-\epsilon_P}{\epsilon_P}\right]_{+}\\
       &+\left[\frac{\EOpp(z)-d_{\max}}{d_{\max}}\right]_{+}\\
       &+\mathbf{1}_{S}\left[\frac{S(\rho)/S(\rho_0)-s_{\max}}{s_{\max}}\right]_{+},
 \end{split}
 \label{eq:violation}
\end{equation}
where $[t]_{+}=\max(0,t)$ and $\mathbf{1}_{S}$ enables the optional storage bound.
All denominators are positive in the experimental requests.
The diagnostic model carries an infeasible status, and the reported component violations show which requirements remain unsatisfied.
A change to a bound requires a user instruction.

\section{Experiments and Results}\label{sec:experiments}

\subsection{Experiment Setup}

\subsubsection{Dataset and Backbone}

The experiments are conducted on Fitzpatrick-17k~\cite{groh2021evaluating}, a clinical image dataset for classifying 114 dermatological conditions.
Following previous works, we group Fitzpatrick types 1--3 as Light and types 4--6 as Dark, using skin tone as the sensitive attribute.

We use a pre-trained VGG-11~\cite{simonyan2014very} adapted to $112 \times 112$ inputs and 114 outputs.
The supplied checkpoint is fixed across experiments and is denoted as Vanilla.
Evaluation uses RGB conversion, a shorter-side resize to 128 pixels, and a center crop, with pixel values scaled to $[0,1]$.
Fine-tuning uses random cropping and horizontal flipping.

\subsubsection{Compression Configurations}

The pre-evaluated candidate set contains $5 \times 2 \times 2 \times 2 = 40$ configurations: five operator sequences, two layer sets, two strengths, and $\beta \in \{5/9,1\}$.
Layer set A contains FC-2, the $4096 \times 4096$ hidden classifier layer; set B contains both FC-1 and FC-2, with FC-1 having $4608 \times 4096$ weights in input-by-output coordinates.
The output classifier layer remains fixed.
Mild and strong settings use pruning or quantization fractions of 0.10 and 0.30, or retained ranks of 1024 and 512 for $L$, respectively.
For both ranks, FairLRF sparsifies 20\% of factor rows or columns and retains half of their rank coordinates.

Scoring uses 256 fixed, class-covering training samples per group and signed diagonal Hessians of the mean cross-entropy loss.
All configurations receive two fine-tuning epochs.
For $P$ or $Q$, the target fraction is reached in two equal increments, each followed by one epoch.
For $L$, both epochs follow factorization and sparsification; each stage of a two-operator configuration receives one epoch.
This shared fine-tuning budget provides a common basis for comparing configurations.

\subsubsection{User Requirements and Search Methods}

The preferred trade-off among accuracy, fairness, and efficiency can be significantly different across various situations.
For simplicity, we consider two straightforward requests fixed before search.
Request U permits a maximum $\Pavg$ loss of one percentage point, requires EOpp no worse than Vanilla, and has no storage upper bound.
Request F permits an average precision loss of up to three percentage points, requires $EOpp \leq 0.20$, and limits storage to 80\% of Vanilla.
Both requests are applied to validation measurements during search.
Feasibility checks use full-precision measurements.

To compare with FCA, we adopt random search~\cite{bergstra2012random}, a tree-structured Parzen estimator (TPE)~\cite{bergstra2011algorithms}, and one-shot LLM planning as baseline methods, together with Vanilla as their common starting point.
TPE uses a constrained multivariate categorical model with exact acquisition maximization over the unobserved catalogue.
Each search method receives the same three initial observations, one mild configuration from each individual operator family.
These observations count toward the limits of 12 distinct candidates and 120 model minutes per session.
Random search samples without replacement.
One-shot planning commits to nine further candidates before receiving their outcomes, while FCA adapts subsequent choices and can stop early.
We conduct three runs per search method under each request, yielding 24 search trajectories.

Both language-model search methods use GLM-5.3-Flash with the same generation settings and per-response allowance.
All search methods query the same pre-evaluated candidate set, with unqueried results kept hidden.
Accumulated model evaluation time includes the queried configuration's preparation, scoring, fine-tuning, validation, and I/O.
Language-model request time is reported separately.
The main comparison combines the completed LLM runs with the corresponding Random/TPE reference trajectories.

\begin{figure}[t]
\centering
\includegraphics[width=0.92\columnwidth]{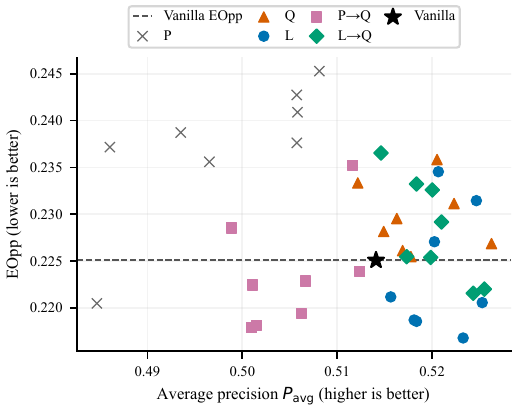}
\caption{Validation outcomes of all 40 compression configurations.
Lower-right points have higher precision and lower EOpp.
The dashed line is the Vanilla EOpp upper bound used by U.
}
\label{fig:candidates}
\end{figure}

\begin{table}[t]
\caption{
Within-family selection under U on validation.
Each compression family contains eight configurations.
Setting gives layer set, strength (mild or strong), and $\beta$.
Selection follows the common rule of minimum storage among feasible observations.
N/A denotes not applicable to Vanilla; \texttt{--} indicates that no feasible configuration is available.
}
\label{tab:operators}
\centering
% \small
% \setlength{\tabcolsep}{3pt}
\begin{tabular}{@{}lclrrr@{}}
\toprule
Method & Feas. & Setting & $\Pavg\uparrow$ & EOpp$\downarrow$ & $S/S_0\downarrow$\\
\midrule
Vanilla & N/A & N/A & 0.5141 & 0.2251 & 1.0000\\
$P$ & 0/8 & -- & -- & -- & -- \\
$Q$ & 0/8 & -- & -- & -- & -- \\
$L$ & 5/8 & B/s/1 & 0.5233 & 0.2168 & 0.4046 \\
$P\!\rightarrow\!Q$ & 3/8 & A/m/1 & 0.5124 & 0.2239 & 1.0000 \\
$L\!\rightarrow\!Q$ & 2/8 & B/m/$5/9$ & 0.5243 & 0.2216 & 0.5953 \\
\bottomrule
\end{tabular}

\end{table}

\subsection{Results of the Compression Operators}

Figure~\ref{fig:candidates} shows that the operator and its configuration both influence the resulting trade-off.
Among the 40 configurations, 10 satisfy U: five from $L$, three from $P\!\rightarrow\!Q$, and two from $L\!\rightarrow\!Q$.
The individual $P$ and $Q$ configurations fall outside the feasible region under the evaluated fine-tuning budget.
Feasible configurations therefore span several families and require evaluation at the configuration level.

Table~\ref{tab:operators} reports the best U-feasible configuration within each family.
The best candidate in the set is $L$ on both hidden classifier layers with rank 512 and $\beta=1$.
It has $\Pavg = 0.5233$, $EOpp = 0.2168$, and 73,385,928 inference tensor bytes, corresponding to a 59.54\% storage reduction from Vanilla.
Validation precision increases by 0.918 percentage points and EOpp decreases by 0.00829.
These measurements describe the complete configuration after compression and fine-tuning.

The composition results illustrate the need for feedback.
Adding quantization to this $L$ configuration gives $\Pavg=0.5146$ and EOpp $=0.2365$ at the same tensor storage, placing the resulting $L\!\rightarrow\!Q$ candidate outside U.
Meanwhile, $P\!\rightarrow\!Q$ supplies three feasible configurations even though the evaluated individual $P$ and $Q$ families supply none.
Thus, both the operator sequence and its settings affect the attainable accuracy--fairness trade-off.
FCA evaluates the complete configuration before accepting it under the user's constraints.
The storage ratios of $P$, $Q$, and $P\!\rightarrow\!Q$ remain 1.0 because their inference tensors are stored densely in FP32.

\subsection{Comparison of Search Methods}

\begin{table*}[t]
\caption{
Replay of search methods on the same pre-evaluated candidate set, with three runs per row.
Storage, precision, and EOpp are coordinate-wise medians over feasible final selections; $v$ is the median normalized violation over all runs.
Queries and model minutes are means.
All 24 trajectories in the main comparison complete.
}
\label{tab:controllers}
\centering
% \small
\begin{tabular}{@{}llcrrrrrr@{}}
\toprule
Request & Method & Feasible & $S/S_0\downarrow$ & $\Pavg\uparrow$ & EOpp$\downarrow$ & $v\downarrow$ & Queries & Model min\\
\midrule
U & Random & 3/3 & 0.5953 & 0.5243 & 0.2212 & 0.0000 & 12.00 & 21.91 \\
U & TPE & 3/3 & 0.5953 & 0.5233 & 0.2186 & 0.0000 & 12.00 & 21.78 \\
U & One-shot & 3/3 & 0.4046 & 0.5233 & 0.2168 & 0.0000 & 12.00 & 22.11 \\
U & FCA & 3/3 & 0.4046 & 0.5233 & 0.2168 & 0.0000 & 7.33 & 13.66 \\
\midrule
F & Random & 0/3 & -- & -- & -- & 0.1059 & 12.00 & 21.91 \\
F & TPE & 0/3 & -- & -- & -- & 0.0839 & 12.00 & 21.93 \\
F & One-shot & 0/3 & -- & -- & -- & 0.0839 & 12.00 & 22.13 \\
F & FCA & 0/3 & -- & -- & -- & 0.0839 & 10.67 & 19.53 \\
\bottomrule
\end{tabular}

\end{table*}

\begin{figure*}[t]
\centering
\includegraphics[width=0.91\textwidth]{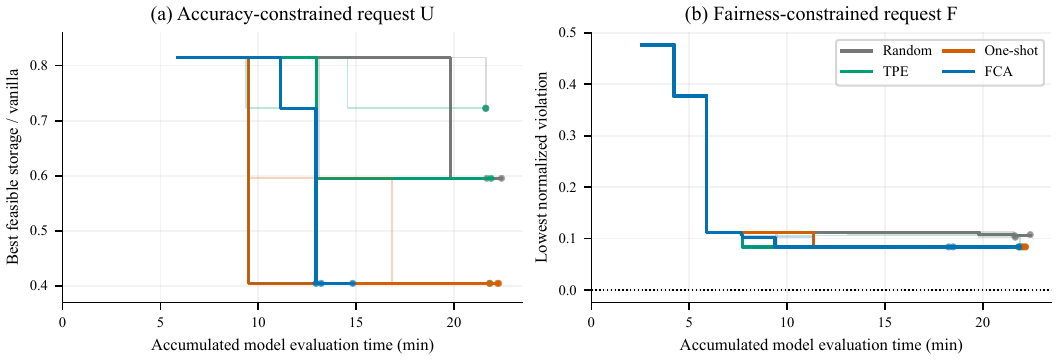}
\caption{
Search progress under U and F.
Thin lines show individual runs and thick lines show point-wise medians;
endpoint dots mark stopping times.
The last observation is carried forward for the median after a session stops.
U shows feasible storage and F shows the lowest observed violation.
The horizontal axis reports the accumulated recorded cost of queried model evaluations;
language-model request time is reported separately.
}
\label{fig:budget}
\end{figure*}

Table~\ref{tab:controllers} compares search method outcomes.
Under U, both one-shot planning and FCA select the best feasible configuration in the candidate set in all three runs, with median storage ratio 0.4046.
Random and TPE have median storage ratio 0.5953.
The shared initialization already includes a U-feasible low-rank candidate, so this comparison measures improvement in storage among feasible selections.
The model selected by FCA requires 32.04\% less storage than the median storage requirement of models selected by Random or TPE.

FCA evaluates 7.33 candidates on average under U, compared with 12 for one-shot planning, with corresponding accumulated model evaluation times of 13.66 and 22.11 minutes.
Its adaptive stopping policy saves evaluations once it has identified a satisfactory trade-off.
One-shot planning executes its committed list, so the comparison reflects the complete policies, including their different stopping rules.

Under F, the lowest EOpp in the entire candidate set is 0.2168, above the upper bound of 0.20.
All search methods therefore report an unmet request.
FCA, one-shot planning, and TPE reach the same median violation of 0.0839, compared with 0.1059 for Random.
The 0.8 storage bound also rules out dense $P$, $Q$, and $P\!\rightarrow\!Q$ representations from their known storage requirements.
This case illustrates how structural information guides the search while measured group performance determines the remaining feasibility gap.
Figure~\ref{fig:budget} shows the evolution of the selected storage and violation as observations accumulate.

\begin{table}[t]
\caption{
Held-out results for three distinct compression configurations selected before test evaluation.
}
\label{tab:test}
\centering
% \small
\begin{tabular}{@{}lrrrr@{}}
\toprule
Configuration: layers/strength/$\beta$ & $S/S_0\downarrow$ & $\Pavg\uparrow$ & EOpp$\downarrow$\\
\midrule
Vanilla & 1.0000 & 0.5275 & 0.2650 \\
$L$: A/s/$5/9$ & 0.7225 & 0.5491 & 0.2641 \\
$L$: B/m/1 & 0.5953 & 0.5355 & 0.2636 \\
$L$: B/s/1 & 0.4046 & 0.5357 & 0.2562 \\
\bottomrule
\end{tabular}

\end{table}

\subsection{Repeated Fine-Tuning and Held-Out Testing}

We evaluate three distinct compression configurations selected from a designated search run and fixed before test evaluation.
Each configuration is compressed and evaluated from the same checkpoint.
The selected configurations are $L$ with A/strong/$5/9$, B/mild/1, and B/strong/1.
Matching existing measurements for these configurations are reused.

Table~\ref{tab:test} reports held-out performance.
For B/strong/1, the average precision on the test set is 0.5357 compared with 0.5275 for Vanilla, while EOpp is 0.2562 compared with 0.2650 for Vanilla.
Meanwhile, it reaches 59.54\% storage reduction.

\subsection{Planning Cost}

\begin{table}[t]
\caption{Language-model cost summed over three runs per row.
Completion tokens include reported reasoning tokens.
API minutes measure request durations and exclude model execution and the separate online cases.}
\label{tab:api}
\centering
% \small
\begin{tabular}{@{}llrrr@{}}
\toprule
Request & Method & Requests & Completion tokens & API min\\
\midrule
U & One-shot & 3 & 69,653 & 18.82 \\
U & FCA & 22 & 75,421 & 15.80 \\
F & One-shot & 3 & 67,774 & 19.10 \\
F & FCA & 35 & 154,014 & 38.65 \\
\bottomrule
\end{tabular}

\end{table}

Table~\ref{tab:api} reports the computational cost of planning.
Under U, FCA uses 22 requests and 75,421 completion tokens across the three runs, compared with three requests and 69,653 tokens for one-shot planning.
The accumulated request durations are 15.80 and 18.82 minutes, respectively.
Together with the candidate counts in Table~\ref{tab:controllers}, these results describe the cost of producing the same final U selection.

The stricter F request requires more interaction.
FCA uses 35 requests and 154,014 completion tokens, compared with three requests and 67,774 tokens for one-shot planning, and both reach the same least-violating model.
The separate request and model-time measurements characterize the cost of the complete search workflow.

\subsection{Online Execution and User Feedback}

Two online cases evaluate fresh compression configurations under a six-candidate limit per case.
In both cases, immediate reselection from existing observations is verified before further model evaluations.

Under U, the user increases the permitted average precision loss from one to two percentage points.
The final selection is $L$ with A/strong/$5/9$, yielding validation $\Pavg=0.5184$, EOpp $=0.2186$, and storage ratio 0.7225.

Under F, the user tightens the EOpp upper bound from 0.20 to 0.19.
FCA reports $L$ with B/strong/1 as infeasible, with normalized violation 0.1410.
The report identifies EOpp as the unsatisfied constraint.

\section{Discussion}\label{sec:discussion}

\subsection{Implications for Compression Selection}

The composition results show that an additional operator can change the accuracy--fairness trade-off of an existing compression configuration.
This makes the complete configuration the appropriate unit of comparison for compression selection.
The U and F results further illustrate how the requested trade-off determines feasibility: a model accepted under U can remain infeasible under F despite its storage reduction.
FCA makes these differences explicit by associating each recommendation with the measured metrics and the constraints used for selection.

\subsection{Representation-Aware Design}

The FPGA considerations in Section~\ref{sec:introduction} explain why compression should be organized around the resulting representation.
Pruning changes connectivity, quantization constrains arithmetic values, and factorization replaces a layer with smaller matrix operations.
Their suitability depends on the resources and execution patterns supported by the target design.
FCA expresses these distinctions through operator descriptions and compatible operator sequences, and uses model-level constraints to preserve the user's accuracy and fairness requirements during selection.
Following the analysis in~\cite{guo2024hardware}, such organization makes deployment assumptions explicit and supports extending fairness evaluation to target hardware.

\section{Conclusion}

In this paper, we proposed FairCompressAgent, an agentic framework for selecting fairness-aware compression configurations under user-defined requirements.
FCA combines pruning, quantization, and sparse low-rank factorization through a common operator interface and connects planning to measured model transformations.
Experiments on Fitzpatrick-17k demonstrate a 59.54\% reduction in inference tensor storage for the model selected on validation under the accuracy-constrained request, together with explicit reporting of unmet constraints and online adaptation to revised requirements.
The shared evaluation record connects these selections to reproducible model execution and supports continued refinement as the user's requirements evolve.

\bibliographystyle{IEEEtran}
\bibliography{references}

\end{document}